\documentstyle[12pt,mkstyle,twoside,multicol]{article}
\title{Too Fast Causal Inference under Causal 
Insufficiency
} 
\shorttitle{Too
Fast Causal Inference}

\newcommand{\Bem}[1]{}

\renewcommand{\baselinestretch}{1.6}

\newcommand{\Speicher}{}
\newcommand{\figopen}[2]{    
\renewcommand{\Speicher}{\caption{#1} \label{#2}}
 \begin{figure} 
\begin{center}
}
\newcommand{\figclose}{\end{center}
\Speicher
 \end{figure} }

\newcommand{\Bilder}[1]{#1}
\newcommand{\dasBild}[1]{#1 \renewcommand{#1}{}  }

\Bilder{\input TO_PIC.INI}

\date{}
\begin{document}

\machetitel







\begin{center}
ABSTRACT 
\end{center}

{
\renewcommand{\baselinestretch}{0.8}

{
Causally insufficient structures (models with latent or hidden variables, or 
with confounding etc.) 
of joint probability distributions 
have been subject of intense study not only in 
statistics, but also in various AI systems. 
 In AI, belief networks, being 
representations of joint probability distribution with an underlying directed 
acyclic graph structure, are paid special attention due to the fact that 
efficient reasoning (uncertainty propagation) methods have been developed for 
 belief network structures. Algorithms have been therefore developed to 
acquire 
the belief network structure from data. As artifacts due 
to variable hiding negatively influence the performance of derived belief 
networks, models with latent variables have been studied and several 
algorithms for learning belief network structure under causal insufficiency 
have also been developed. 
Regrettably, some of them are known already to be 
erroneous (e.g. IC algorithm of \cite{Pearl:Verma:91}). This paper is devoted 
to another algorithm, the Fast Causal Inference (FCI)  Algorithm of 
\cite{Spirtes:93}. It is proven by a specially constructed example that this 
algorithm, as it stands in \cite{Spirtes:93}, 
 is also erroneous. Fundamental reason for failure of this algorithm 
is the temporary introduction of non-real links between nodes of the network 
with the intention of later removal. While for trivial dependency structures 
these non-real links may be actually removed, this may not be the case for 
complex ones, e.g. for the case described in this paper. A remedy 
of this failure is proposed.\\
} 

}

\newpage

\section{Introduction}

Various expert systems, dealing with uncertain data and knowledge, 
possess 
knowledge representation in terms of a 
belief network (e.g. knowledge base of the MUNIM system 
\cite{Andreassen:87} , ALARM network 
\cite{Cooper:92} etc.).
A 
number of  efficient algorithms for propagation of uncertainty within 
belief networks and their derivatives 
 have been developed,  compare e.g.
 \cite{Pearl:88,Shachter:90b,Shenoy:90}. 

Belief networks, causal networks, or influence diagrams, are terms frequently 
used interchangeably. They are quite 
popular for expressing causal relations under multiple variable setting 
both for deterministic and non-deterministic (e.g. stochastic) relationships 
in various 
domains: statistics, philosophy, artificial intelligence 
\cite{Geiger:90,Spirtes:90b}. Though a belief network (a
representation of the joint probability 
distribution, see \cite{Geiger:90}) and a causal network (a representation 
of 
causal relationships \cite{Spirtes:90b}) are intended to mean different 
things, 
they are closely related. Both assume an underlying dag (directed acyclic 
graph) structure of relations among variables and if Markov condition and 
faithfulness condition \cite{Spirtes:93} are met, then a causal network is in 
fact a belief network. The difference comes to appearance when we recover 
belief network and causal network structure from data. A dag of a belief 
network is satisfactory if the generated probability distribution fits the 
data, may be some sort of minimality is required. A causal network structure 
may be impossible to recover completely from data as not all directions of 
causal links may be uniquely determined \cite{Spirtes:93}. Fortunately, if  
we deal with causally sufficient sets of variables (that is whenever 
  significant influence variables are not omitted from observation), then 
there 
exists the possibility to identify the family of belief networks a causal 
network belongs to \cite{Verma:Pearl:90} (see also \cite{Klopotek:93p5}). 

Regrettably, to our knowledge, a similar result is not directly known for 
causally insufficient sets of variables (that is when significant influence 
variables are hidden) - "Statistical indistinguishability is less well 
understood when graphs can contain variables representing unmeasured common 
causes" (\cite{Spirtes:93}, p. 88). Latent (hidden) variable 
identification has been investigated intensely both for belief networks (e.g. 
\cite{Pearl:86h,Golmard:Mallet:89,Liu:Wilkins:Ying:Bian:90,Cooper:92}) and
causal networks 
(
\cite{Pearl:Verma:91,Spirtes:90b,Spirtes:93,Glymour:87,Glymour:Spirtes:88}),
 beside the immense research 
effort in traditional statistics 
(to mention results of Spearman on vanishing tetrad differences from the
beginning of this century to recent LISREL and EQS techniques
- see \cite{Spirtes:90c} for a comparative study of these techniques with 
causal network approaches in AI).
The algorithm of \cite{Cooper:92} recovers the most probable location of a 
hidden variable. 
 Whereas the CI algorithm of \cite{Spirtes:93} recovers exact 
locations of common causes, but clearly not all of them. In fact, the CI 
algorithm does not provide a dag, but rather a graph with edges fully 
(unidirected or bidirected) or 
partially oriented, or totally non-oriented with additional constraints for 
 edge directions at other edges.  Partially or non-oriented edges  may prove 
to 
be either directed or bidirected edges. 
Alternatively, the IC algorithm of   Pearl and Verma \cite{Pearl:Verma:91} 
 tried to recover the family of "minimal latent models" (a family of dags 
close 
to the data), but, as Spirtes et al. claim in \cite{Spirtes:93}, page 200,  
 "Unfortunately, the 
two main claims about the output of the Inductive Causation Algorithm made in 
the paper ... are false.". 
Hence the
big question  is whether or not the bidirectional edges (that is 
 indications of a common cause) are the only ones necessary to develop a 
belief 
network out of the product of CI, or must there be some other hidden 
variables added (e.g. by  guessing).  
We answer this question in favour of the CI algorithm elsewhere 
\cite{Klopotek:93g}. However, as Spirtes et al. state, their CI 
algorithm 
is feasible only for a small number of variables and hence they developed an 
 "accelerated" version of the CI algorithm: the FCI algorithm, which also has 
a 
partial including path graph as its output. The question formulated for 
CI needs thus to be repeated for the FCI algorithm. Regrettably as it 
is, the FCI algorithm, as it stands in \cite{Spirtes:93}, 
 cannot be accommodated for recovery of possible 
belief networks as it introduces into the causal structure causal arrows  
which are not actually present in the data, and due to this fact a resulting 
 belief network would contain dependencies not present in the data, but what 
is 
worse, it would exhibit independencies not present in the data. 

We sought to recover from this error. First of all, we noticed the discrepance
between the notion of D-SEP and Possible-D-SEP of the FCI algorithm (they
do not agree for a fully oriented including path graph). An attempt to
reconcile these notions proved to be misleading, because while providing
remedy for the first example it lead to errors in a more complex example. 

Therefore a re-elaboration of two stages of FCI is proposed in order to
stabilize the dynamics of Possible-D-SEP under edge removal. \\


\section{Fast Causal Inference Algorithm of Spirtes et al.}

To make this paper self-contained, 
we below remind the Causal Inference (CI) and the Fast Causal Inference (FCI) 
algorithms of Spirtes, Glymour and 
Scheines \cite{Spirtes:93} together with some basic notation used therein. 
Recalling CI algorithm is necessary as FCI refers to CI in its final phase.
The text of this section is to a large extent a citation from 
\cite{Spirtes:93},  and  quotation  marks  will  be  dropped   for 
readability.
 
Essentially, the CI algorithm recovers partially the structure of 
an including 
path graph. Given a directed acyclic graph G with the set of hidden nodes  
$V_h$ 
and visible nodes $V_s$ representing a causal network CN, an including path 
between nodes A and  B belonging to $V_s$ is a path in the graph G such that  
the only visible nodes (except for A and B) on the path are those where edges 
of the path meet head-to-head and there exists a directed  path  in G from 
such a node 
to either A or B. An including path graph for G is such a graph over $V_s$ in 
which if nodes A and B are connected by an including path in G ingoing into A 
and B, then A and B are connected by a bidirectional edge $A<->B$. Otherwise 
if they are connected by an including path in G outgoing from A and ingoing 
into B then A and B are connected by an unidirectional edge $A->B$.  

A partially oriented including path graph contains the following types of 
edges unidirectional: $A->B$, bidirectional $A<->B$, partially oriented 
$Ao->B$ and non-oriented $Ao-oB$, as well as some local constraint 
information $A*-\underline{*B*}-*C$ 
 meaning that edges between A and B and 
between B and C cannot meet head to head at B. (Subsequently an asterisk  
($*$) 
means any orientation of an edge end: e.g. $A*->B$ means either $A->B$ or 
$Ao->B$ or $A<->B$).

 In a partially oriented including graph $\pi$ (see \cite{Spirtes:93}, pp.: 
181-182)\\
\begin{itemize}
\item[(i)] A is a parent of B if and only if $A->B$ in $\pi$.
\item[(ii)] B is a collider along the path $<A,B,C>$ if and only if 
$A*->B<-*C$ in $\pi$.\\
B is a definite non-collider on undirected path U 
if and only if either B is an end-point of U, or there exist vertices A and C
such  that U contains one of the subpaths $A<--B*-*C$, 
 $A*-*B-->C$, or   $A*-\underline{*B*}-*C$,  (see
Glossary of \cite{Spirtes:93}). 
\item[(iii)] An edge between B and A is into A iff $A<-*B$ in $\pi$
\item[(iv)] An edge between B and A is out of A iff $A->B$ in $\pi$.
\item[(v)] A is d-separated from B given set S iff A and B are conditionally 
independent given S.
\item[(vi)] A and B are d-connected given node C iff  there  exists  no 
such set S 
containing C such that A and B are conditionally 
independent given S.
\item[(vii)] In a partially oriented including path graph $\pi'$, U is a 
definite 
 discriminating path for B if and only if U is an undirected path between X 
and 
Y containing B, $B \neq X, B \neq Y$, every vertex on U except for B and the 
endpoints is a collider or a definite non-collider on U and:\\
(a) if V and V" are adjacent on U, and V" is between V and B on U, then 
$V*->V"$ on U,\\
(b) if V is between X and B on U and V is a collider on U, then $V->Y$ in 
$\pi$, else $V<-*Y$ on $\pi$\\
(c) if V is between Y and B on U and V is a collider on U, then $V->X$ in 
$\pi$, else $V<-*X$ on $\pi$\\
(d) X and Y are not adjacent in $\pi$.\\
\item[viii)] U is a directed path  from X to Y iff there exists an undirected
path between X and Y such that if V is adjacent to X on U then 
$X->V$ in $\pi$, if $V$ is adjacent to Y on V, then $V->Y$, if V and V" are 
adjacent on U 
and V is between X and V" on U, then $V->V"$ in $\pi$.
\end{itemize}%

{\noindent \bf The Causal Inference (CI) Algorithm:} (see \cite{Spirtes:93}, 
pp.: 183)\\
Input: Empirical joint probability distribution\\
Output: partial including path graph $\pi$.\\
\begin{description}
\item[A)] Form the complete undirected graph Q on the vertex set V.
\item[B)] if A and B are d-separated given any subset S of V, remove the edge 
between 
A and B, and record S in Sepset(A,B) and Sepset(B,A). 
\item[C)] Let F be the graph resulting from step B). Orient each edge 
o-o.  For each 
triple of vertices A,B,C such that the pair A,B and the pair B,C are each 
adjacent in F, but the pair A,C are not adjacent in F, orient  $A*-*B*-*C$ 
as 
$A*->B<-*C$ if and only if B is not in Sepset(A,C), and orient 
 $A*-*B*-*C$ 
as $A*-\underline{*B*}-*C$ if and only if B is in Sepset(A,C).
\item[D)] Repeat
\begin{itemize}
\item[if] there is a directed path from A to B, and an edge  $A*-*B$, orient 
 $A*-*B$
as $A*->B$,
\item[else if] B is a collider along $<A,B,C>$ in $\pi$, B is adjacent to D, 
and A and C are not d-connected given D, then orient $B*-*D$ as 
$B<-*D$ ,
\item[else if] U is a definite discriminating path between A and B for M in 
$\pi$ and 
P and R are adjacent to M on U, and P-M-R is a triangle, then\\
if M is in Sepset(A,B) then M is marked as non-collider on subpath 
$P*-\underline{*M*}-R$\\
else $P*-*AM*-*R$ is oriented  as $P*->M<-*R$,
\item[else if] $P*-\underline{>M*}-*R$ then orient  as $P*->M->R$.
\item[until] no more edges can be oriented.
\end{itemize}
\item[End of CI]
\end{description}
 
To understand the proper FCI algorithm, some additional definitions are 
necessary:
 
\begin{itemize}
\item[ix)] 
 In a full including graph $\pi_0$ 
 V is in D-Sep(A,B) iff $V\neq A$ and there is an undirected path from V to  
A such 
 that all the nodes on the path are colliders having either A or B as their 
definite successor. (see \cite{Spirtes:93}, p. 
187)
\item[x)] 
"For a given partially constructed partially oriented including path graph
$\pi$, {\bf Possible-D-Sep}(A,B) is defined as follows: 
If $A\neq B$, V is in  {\bf Possible-D-Sep}(A,B) in $\pi$ if and only if
$V\neq A$, and there is an undirected path U between A and V in $\pi$ such
that for every subpath $<X,Y,Z>$ of U either Y is a collider on the subpath,
or Y is not a definite non-collider and on U, and X, Y, and Z form a triangle
in $\pi$. " (\cite{Spirtes:93}, p.187 below Fig.18, repeated in Glossary
therein). 
\end{itemize}

{\noindent \bf The Fast Causal Inference (FCI) Algorithm:} (see 
\cite{Spirtes:93}, p.: 188)\\
Input: Empirical joint  distribution\\
Output: partial including path graph $\pi$.\\
\begin{description}
\item[A)] Form the complete undirected graph Q on the vertex set V.
\item[B)]  n = 0;\\
repeat\\
repeat\\
select an ordered pair of variables X and Y that are adjacent in Q such that 
Adjacencies(Q,X)-\{Y\} has cardinality greater or equal to n, and a subset S 
of Adjacencies(Q,X)-\{Y\} of cardinality n, and if X and Y are independent 
    given S delete edge between X and Y from Q, and record S in Sepset(X,Y) 
and in 
SepSet(Y,X)\\
until all ordered pairs of adjacent variables such that 
Adjacencies(Q,X)-\{Y\} has cardinality greater than or equal to n and all 
subsets S of Adjacencies(Q,X)-\{Y\} of cardinality n have been tested for 
making X,Y independent.;\\
n=n+1;\\
until for each ordered pair of adjacent vertices X,Y, Adjacencies(Q,X)-\{Y\} 
is of cardinality less than n.
\item[C)] Let F be the undirected graph resulting from step B). Orient each 
edge as $o-o$. For each triple of vertices A,B,C such that the pair A,B and 
the pair B,C are each adjacent in F, but the pair A,C are not adjacent in F, 
orient $A*-*B*-*C$ as $A*->B<-*C$ if and only if B is not in Sepset(A,C).
\item[D)] For each pair of variables A and B adjacent in F', if A and B are 
independent given any subset S of Possible-D-SEP(A,B)-\{A,B\} or any subset S 
of  Possible-D-SEP(B,A)-\{A,B\} in F remove edge between A and B and record S 
in Sepset(A,B) and Sepset(B,A). \\
\item[E)] Reset all edge orientations as $o-o$ and carry out steps C) and D) 
of the Ci algorithm\\

\item[End of FCI]
\end{description}

\section{The Claim of this Paper About FCI}

Let us imagine that we have obtained a partial including path graph from FCI, 
and we want to find a Belief Network representing the joint probability 
distribution out of it. Let us consider the following algorithm:\\

\noindent
{\bf FCI-to-BN Algorithm}\\
Input: Result of the FCI algorithm (a partial including path graph)\\
Output: A belief network
\begin{description}
\item[A)] Accept unidirectional and bidirectional edges obtained from CI.
\item[B)] Orient every edge $Ao->B$ as $A->B$.
\item[C)]  Orient edges of type $Ao-oB$ either as $A<-B$ or $A->B$ so as not 
to 
violate  $P*-\underline{*M*}-*R$ constraints.
\item[End of CI-to-BN]
\end{description}

We claim that:
\begin{th}
The belief network obtained via FCI-to-BN algorithm does not in general keeps 
all the dependencies 
and independencies of the original underlying including path graph.
\end{th}

The rest of this paper   provides a sketchy proof of the above theorem by an 
example. First we demonstrate, that step C) of FCI generates arrow 
 orientations contradicting the edge orientation of the original including 
path 
graph. Then we show that this leads to violation of dependence/independence 
relation in the resulting belief network.
We then suspect that definition (x) above of Possible-D-SEP is not correct
and check another meaning thereof. Though it provides a recovery from the
failure of the initial example, it runs into error on a larger
example.  Finally, 
we rewrite FCI
algorithm altogether to ensure in step D removal of superfluous edges left in
step B. 

\section{FCI, As It Stands, Fails}

Please compare first definitions (ii) of definite non-collider and (x) of 
Possible-D-Sep with the contents of proper FCI algorithm. 
Node Y from definition (x) is not the endpoint on U, proper FCI algorithm
introduces neither unidirectional edges nor $X*-\underline{*Y*}-*Z$
constraints. Hence the phrase "Y is not a definite non-collider" in definition
(x) is absolutely pointless as Y is always not a definite non-collider out of
the construction of FCI algorithm, as it stands in \cite{Spirtes:93}.
We rewrite definition (x) as:
 
\begin{itemize}
\item[x')] 
For a given partially constructed partially oriented including path graph
$\pi$, {\bf Possible-D-Sep}(A,B) is defined as follows: 
If $A\neq B$, V is in  {\bf Possible-D-Sep}(A,B) in $\pi$ if and only if
$V\neq A$, and there is an undirected path U between A and V in $\pi$ such
that for every subpath $<X,Y,Z>$ of U either Y is a collider on the subpath,
or Y is on U, and X, Y, and Z form a triangle
in $\pi$.  
\end{itemize}

 \Bilder{\input FCI.PIC}

\dasBild{\AbbEins}

Let us study a run of the FCI algorithm on a set of
visible (observable) variables
with intrinsic causal relationships from 
Fig.\ref{abbeins}. The double arrows $A<->B$ in this figure are
to be interpreted as follows: there exists a (hidden, not observable) variable
$H_{A,B}$ such that the causal relationship is in fact as follows: $A<-H_{A,
B}->B$. \\ 

The detailed list of nodes and edges is given below:

{\footnotesize 
\begin{multicols}{2}
\begin{verbatim}

NODES

Sub-Network 1

Name  TeX-Name  X-  Y- Coordinates
 ~nZ1 Z_1       20  110
 ~nT1 T_1       20  190
 ~nV1 V_1        0  150
 ~nb1 B_1       40  126
 ~nc1 C_1       70  100

Sub-Network 2

 ~nZ2 Z_2       20  210
 ~nT2 T_2       20  290
 ~nV2 V_2        0  250
 ~nb2 B_2       40  200
 ~nc2 C_2       70  200

Sub-Network 3

 ~nY3 Y_3      200   90
 ~nX3 X_3      200  190
 ~nZ3 Z_3      120  110
 ~nT3 T_3      120  190
 ~nV3 V_3      100  130
 ~nW3 W_3      100  150
 ~nS3 S_3      160  179
 ~nR3 R_3      160  160

 ~nB3 B_3      140  100
 ~nC3 C_3      170  100
 ~nb3 B_3'     140  199
 ~nc3 C_3'     170  199

 ~nP3 P_3      250  100
 ~nL1 L_1      210  120
 ~nL2 L_2      210  130

EDGES

Sub-Network 1    
 ~eZ1<->V3   
 ~eT1<->R3   
 ~eV1-->Z1 ~eV1-->T1 
 ~eZ1-->b1  ~eb1-->R3
 ~eT1-->c1  ~ec1-->V3

Sub-Network 2    
 ~eZ2<->W3   
 ~eT2<->S3   
 ~eV2-->Z2 ~eV2-->T2 
 ~eZ2-->b2  ~eb2-->S3
 ~eT2-->c2  ~ec2-->W3

Sub-Network 3    
 ~eZ3<->R3 ~eR3<->X3 ~eR3-->Y3  
 ~eT3<->S3 ~eS3<->Y3 ~eS3-->X3  ~eX3-->P3
 ~eR3-->P3
 ~eL1-->P3 ~eL2-->P3 
 ~eS3-->L1  ~eS3-->L2
 ~eV3-->L1  ~eV3-->L2
 ~eZ3-->P3


 ~eV3-->Z3 ~eV3-->W3  ~eW3-->T3 
 ~eZ3-->B3 ~eB3-->C3 ~eC3-->Y3
 ~eT3-->b3 ~eb3-->c3 ~ec3-->Y3

\end{verbatim}
\end{multicols}
}

 Step A) of FCI is trivial. Let us consider step B).
We start with n = 0. 

{\footnotesize 
\begin{multicols}{2}
\begin{verbatim}
 FCI stage B, n=0 protocol 
Edge removal:  ~eB3   S3 SepSet {}
Edge removal:  ~eB3   T2 SepSet {}
Edge removal:  ~eB3   V2 SepSet {}
Edge removal:  ~eB3   X3 SepSet {}
Edge removal:  ~eB3   Z2 SepSet {}
Edge removal:  ~eB3   b2 SepSet {}
Edge removal:  ~eB3   c2 SepSet {}
Edge removal:  ~eC3   S3 SepSet {}
Edge removal:  ~eC3   T2 SepSet {}
Edge removal:  ~eC3   V2 SepSet {}
Edge removal:  ~eC3   X3 SepSet {}
Edge removal:  ~eC3   Z2 SepSet {}
Edge removal:  ~eC3   b2 SepSet {}
Edge removal:  ~eC3   c2 SepSet {}
Edge removal:  ~eR3   S3 SepSet {}
Edge removal:  ~eR3   T2 SepSet {}
Edge removal:  ~eR3   V2 SepSet {}
Edge removal:  ~eR3   Z2 SepSet {}
Edge removal:  ~eR3   b2 SepSet {}
Edge removal:  ~eR3   c2 SepSet {}
Edge removal:  ~eS3   T1 SepSet {}
Edge removal:  ~eS3   V1 SepSet {}
Edge removal:  ~eS3   V3 SepSet {}
Edge removal:  ~eS3   Z1 SepSet {}
Edge removal:  ~eS3   Z3 SepSet {}
Edge removal:  ~eS3   b1 SepSet {}
Edge removal:  ~eS3   c1 SepSet {}
Edge removal:  ~eT1   T2 SepSet {}
Edge removal:  ~eT1   V2 SepSet {}
Edge removal:  ~eT1   X3 SepSet {}
Edge removal:  ~eT1   Z2 SepSet {}
Edge removal:  ~eT1   b2 SepSet {}
Edge removal:  ~eT1   c2 SepSet {}
Edge removal:  ~eT2   V1 SepSet {}
Edge removal:  ~eT2   V3 SepSet {}
Edge removal:  ~eT2   Z1 SepSet {}
Edge removal:  ~eT2   Z3 SepSet {}
Edge removal:  ~eT2   b1 SepSet {}
Edge removal:  ~eT2   c1 SepSet {}
Edge removal:  ~eV1   V2 SepSet {}
Edge removal:  ~eV1   X3 SepSet {}
Edge removal:  ~eV1   Z2 SepSet {}
Edge removal:  ~eV1   b2 SepSet {}
Edge removal:  ~eV1   c2 SepSet {}
Edge removal:  ~eV2   V3 SepSet {}
Edge removal:  ~eV2   Z1 SepSet {}
Edge removal:  ~eV2   Z3 SepSet {}
Edge removal:  ~eV2   b1 SepSet {}
Edge removal:  ~eV2   c1 SepSet {}
Edge removal:  ~eV3   X3 SepSet {}
Edge removal:  ~eV3   Z2 SepSet {}
Edge removal:  ~eV3   b2 SepSet {}
Edge removal:  ~eV3   c2 SepSet {}
Edge removal:  ~eX3   Z1 SepSet {}
Edge removal:  ~eX3   Z3 SepSet {}
Edge removal:  ~eX3   b1 SepSet {}
Edge removal:  ~eX3   c1 SepSet {}
Edge removal:  ~eZ1   Z2 SepSet {}
Edge removal:  ~eZ1   b2 SepSet {}
Edge removal:  ~eZ1   c2 SepSet {}
Edge removal:  ~eZ2   Z3 SepSet {}
Edge removal:  ~eZ2   b1 SepSet {}
Edge removal:  ~eZ2   c1 SepSet {}
Edge removal:  ~eZ3   b2 SepSet {}
Edge removal:  ~eZ3   c2 SepSet {}
Edge removal:  ~eb1   b2 SepSet {}
Edge removal:  ~eb1   c2 SepSet {}
Edge removal:  ~eb2   c1 SepSet {}
Edge removal:  ~ec1   c2 SepSet {}

 FCI stage B, n=1 protocol 
Edge removal:  ~eB3   L1 SepSet {V3}
Edge removal:  ~eB3   L2 SepSet {V3}
Edge removal:  ~eB3   P3 SepSet {Z3}
Edge removal:  ~eB3   R3 SepSet {Z3}
Edge removal:  ~eB3   T1 SepSet {V3}
Edge removal:  ~eB3   T3 SepSet {V3}
Edge removal:  ~eB3   V1 SepSet {V3}
Edge removal:  ~eB3   V3 SepSet {Z3}
Edge removal:  ~eB3   W3 SepSet {Z3}
Edge removal:  ~eB3   Z1 SepSet {Z3}
Edge removal:  ~eB3   b1 SepSet {Z3}
Edge removal:  ~eB3   b3 SepSet {Z3}
Edge removal:  ~eB3   c1 SepSet {Z3}
Edge removal:  ~eB3   c3 SepSet {Z3}
Edge removal:  ~eC3   L1 SepSet {B3}
Edge removal:  ~eC3   L2 SepSet {B3}
Edge removal:  ~eC3   P3 SepSet {B3}
Edge removal:  ~eC3   R3 SepSet {B3}
Edge removal:  ~eC3   T1 SepSet {B3}
Edge removal:  ~eC3   T3 SepSet {B3}
Edge removal:  ~eC3   V1 SepSet {B3}
Edge removal:  ~eC3   V3 SepSet {B3}
Edge removal:  ~eC3   W3 SepSet {B3}
Edge removal:  ~eC3   Z1 SepSet {B3}
Edge removal:  ~eC3   Z3 SepSet {B3}
Edge removal:  ~eC3   b1 SepSet {B3}
Edge removal:  ~eC3   b3 SepSet {B3}
Edge removal:  ~eC3   c1 SepSet {B3}
Edge removal:  ~eC3   c3 SepSet {B3}
Edge removal:  ~eL1   R3 SepSet {V3}
Edge removal:  ~eL1   T1 SepSet {V3}
Edge removal:  ~eL1   T2 SepSet {S3}
Edge removal:  ~eL1   V1 SepSet {V3}
Edge removal:  ~eL1   V2 SepSet {S3}
Edge removal:  ~eL1   X3 SepSet {S3}
Edge removal:  ~eL1   Z1 SepSet {V3}
Edge removal:  ~eL1   Z2 SepSet {S3}
Edge removal:  ~eL1   Z3 SepSet {V3}
Edge removal:  ~eL1   b1 SepSet {V3}
Edge removal:  ~eL1   b2 SepSet {S3}
Edge removal:  ~eL1   b3 SepSet {T3}
Edge removal:  ~eL1   c1 SepSet {V3}
Edge removal:  ~eL1   c2 SepSet {S3}
Edge removal:  ~eL1   c3 SepSet {T3}
Edge removal:  ~eL2   R3 SepSet {V3}
Edge removal:  ~eL2   T1 SepSet {V3}
Edge removal:  ~eL2   T2 SepSet {S3}
Edge removal:  ~eL2   V1 SepSet {V3}
Edge removal:  ~eL2   V2 SepSet {S3}
Edge removal:  ~eL2   X3 SepSet {S3}
Edge removal:  ~eL2   Z1 SepSet {V3}
Edge removal:  ~eL2   Z2 SepSet {S3}
Edge removal:  ~eL2   Z3 SepSet {V3}
Edge removal:  ~eL2   b1 SepSet {V3}
Edge removal:  ~eL2   b2 SepSet {S3}
Edge removal:  ~eL2   b3 SepSet {T3}
Edge removal:  ~eL2   c1 SepSet {V3}
Edge removal:  ~eL2   c2 SepSet {S3}
Edge removal:  ~eL2   c3 SepSet {T3}
Edge removal:  ~eP3   T2 SepSet {S3}
Edge removal:  ~eP3   V2 SepSet {S3}
Edge removal:  ~eP3   Z2 SepSet {S3}
Edge removal:  ~eP3   b2 SepSet {S3}
Edge removal:  ~eP3   b3 SepSet {T3}
Edge removal:  ~eP3   c2 SepSet {S3}
Edge removal:  ~eP3   c3 SepSet {T3}
Edge removal:  ~eR3   T3 SepSet {V3}
Edge removal:  ~eR3   V1 SepSet {Z1}
Edge removal:  ~eR3   W3 SepSet {V3}
Edge removal:  ~eR3   Z1 SepSet {b1}
Edge removal:  ~eR3   b3 SepSet {V3}
Edge removal:  ~eR3   c1 SepSet {T1}
Edge removal:  ~eR3   c3 SepSet {V3}
Edge removal:  ~eS3   V2 SepSet {Z2}
Edge removal:  ~eS3   Z2 SepSet {b2}
Edge removal:  ~eS3   b3 SepSet {T3}
Edge removal:  ~eS3   c2 SepSet {T2}
Edge removal:  ~eS3   c3 SepSet {T3}
Edge removal:  ~eT1   T3 SepSet {V3}
Edge removal:  ~eT1   V3 SepSet {c1}
Edge removal:  ~eT1   W3 SepSet {c1}
Edge removal:  ~eT1   Z1 SepSet {V1}
Edge removal:  ~eT1   Z3 SepSet {c1}
Edge removal:  ~eT1   b1 SepSet {V1}
Edge removal:  ~eT1   b3 SepSet {c1}
Edge removal:  ~eT1   c3 SepSet {c1}
Edge removal:  ~eT2   T3 SepSet {W3}
Edge removal:  ~eT2   W3 SepSet {c2}
Edge removal:  ~eT2   X3 SepSet {S3}
Edge removal:  ~eT2   Y3 SepSet {c2}
Edge removal:  ~eT2   Z2 SepSet {V2}
Edge removal:  ~eT2   b2 SepSet {V2}
Edge removal:  ~eT2   b3 SepSet {c2}
Edge removal:  ~eT2   c3 SepSet {c2}
Edge removal:  ~eT3   V1 SepSet {V3}
Edge removal:  ~eT3   V2 SepSet {W3}
Edge removal:  ~eT3   V3 SepSet {W3}
Edge removal:  ~eT3   X3 SepSet {S3}
Edge removal:  ~eT3   Z1 SepSet {W3}
Edge removal:  ~eT3   Z2 SepSet {W3}
Edge removal:  ~eT3   Z3 SepSet {W3}
Edge removal:  ~eT3   b1 SepSet {W3}
Edge removal:  ~eT3   b2 SepSet {W3}
Edge removal:  ~eT3   c1 SepSet {W3}
Edge removal:  ~eT3   c2 SepSet {W3}
Edge removal:  ~eT3   c3 SepSet {b3}
Edge removal:  ~eV1   V3 SepSet {T1}
Edge removal:  ~eV1   W3 SepSet {T1}
Edge removal:  ~eV1   Z3 SepSet {T1}
Edge removal:  ~eV1   b1 SepSet {Z1}
Edge removal:  ~eV1   b3 SepSet {T1}
Edge removal:  ~eV1   c1 SepSet {T1}
Edge removal:  ~eV1   c3 SepSet {T1}
Edge removal:  ~eV2   W3 SepSet {T2}
Edge removal:  ~eV2   X3 SepSet {Z2}
Edge removal:  ~eV2   Y3 SepSet {T2}
Edge removal:  ~eV2   b2 SepSet {Z2}
Edge removal:  ~eV2   b3 SepSet {T2}
Edge removal:  ~eV2   c2 SepSet {T2}
Edge removal:  ~eV2   c3 SepSet {T2}
Edge removal:  ~eV3   b1 SepSet {Z1}
Edge removal:  ~eV3   b3 SepSet {W3}
Edge removal:  ~eV3   c3 SepSet {W3}
Edge removal:  ~eW3   X3 SepSet {S3}
Edge removal:  ~eW3   Z1 SepSet {V3}
Edge removal:  ~eW3   Z3 SepSet {V3}
Edge removal:  ~eW3   b1 SepSet {V3}
Edge removal:  ~eW3   b2 SepSet {Z2}
Edge removal:  ~eW3   b3 SepSet {T3}
Edge removal:  ~eW3   c1 SepSet {V3}
Edge removal:  ~eW3   c3 SepSet {T3}
Edge removal:  ~eX3   Z2 SepSet {S3}
Edge removal:  ~eX3   b2 SepSet {S3}
Edge removal:  ~eX3   b3 SepSet {S3}
Edge removal:  ~eX3   c2 SepSet {S3}
Edge removal:  ~eX3   c3 SepSet {S3}
Edge removal:  ~eY3   b2 SepSet {Z2}
Edge removal:  ~eZ1   Z3 SepSet {V3}
Edge removal:  ~eZ1   b3 SepSet {V3}
Edge removal:  ~eZ1   c1 SepSet {V1}
Edge removal:  ~eZ1   c3 SepSet {V3}
Edge removal:  ~eZ2   b3 SepSet {W3}
Edge removal:  ~eZ2   c2 SepSet {V2}
Edge removal:  ~eZ2   c3 SepSet {W3}
Edge removal:  ~eZ3   b1 SepSet {V3}
Edge removal:  ~eZ3   b3 SepSet {V3}
Edge removal:  ~eZ3   c1 SepSet {V3}
Edge removal:  ~eZ3   c3 SepSet {V3}
Edge removal:  ~eb1   b3 SepSet {Z1}
Edge removal:  ~eb1   c1 SepSet {Z1}
Edge removal:  ~eb1   c3 SepSet {Z1}
Edge removal:  ~eb2   b3 SepSet {Z2}
Edge removal:  ~eb2   c2 SepSet {Z2}
Edge removal:  ~eb2   c3 SepSet {Z2}
Edge removal:  ~eb3   c1 SepSet {T3}
Edge removal:  ~eb3   c2 SepSet {T3}
Edge removal:  ~ec1   c3 SepSet {V3}
Edge removal:  ~ec2   c3 SepSet {W3}

 Number of edges: 73

 FCI stage B, n=2 protocol 
Edge removal:  ~eB3   Y3 SepSet {C3,Z3}
Edge removal:  ~eL1   L2 SepSet {S3,V3}
Edge removal:  ~eL1   T3 SepSet {S3,V3}
Edge removal:  ~eL1   W3 SepSet {V3,S3}
Edge removal:  ~eL1   Y3 SepSet {V3,S3}
Edge removal:  ~eL2   T3 SepSet {S3,V3}
Edge removal:  ~eL2   W3 SepSet {V3,S3}
Edge removal:  ~eL2   Y3 SepSet {V3,S3}
Edge removal:  ~eP3   T3 SepSet {S3,V3}
Edge removal:  ~eP3   W3 SepSet {S3,V3}
Edge removal:  ~eT3   Y3 SepSet {W3,b3}
Edge removal:  ~eW3   Y3 SepSet {T3,V3}
Edge removal:  ~eY3   Z2 SepSet {V3,b3}
Edge removal:  ~eY3   b3 SepSet {V3,c3}
Edge removal:  ~eY3   c2 SepSet {V3,c3}

 Number of edges: 58

 FCI stage B, n=3 protocol 
Edge re.:  ~eV3   Y3 SepSet {R3,W3,Z3}
Edge re.:  ~eT1   Y3 SepSet {R3,C3,c3}
Edge re.:  ~eV1   Y3 SepSet {R3,C3,c3}
Edge re.:  ~eY3   Z1 SepSet {C3,R3,c3}
Edge re.:  ~eY3   Z3 SepSet {C3,R3,c3}
Edge re.:  ~eY3   b1 SepSet {R3,C3,c3}
Edge re.:  ~eY3   c1 SepSet {R3,C3,c3}

 Number of edges: 51

 FCI stage B, n=4 protocol 
Edge re.:  ~eP3   Y3 SepSet {R3,S3,V3,Z3}
Edge re.:  ~eP3   Z1 SepSet {T1,V1,V3,b1}
Edge re.:  ~eP3   c1 SepSet {T1,V1,V3,b1}

 Number of edges: 48

 FCI stage B, n=5 protocol 
Edge re.:  ~eP3   S3 SepSet {L2,R3,L1,X3,Z3}
Edge re.:  ~eP3   T1 SepSet {R3,L1,L2,X3,Z3}
Edge re.:  ~eP3   V1 SepSet {L1,L2,R3,X3,Z3}
Edge re.:  ~eP3   V3 SepSet {L2,R3,L1,X3,Z3}
Edge re.:  ~eP3   b1 SepSet {R3,L1,X3,Z3,L2}

 Number of edges: 43

 FCI stage B, n=6 protocol 
 Number of edges: 43

 FCI stage B, n=7 protocol 
 Number of edges: 43

 FCI stage B, n=8 protocol 
 Number of edges: 43

 FCI stage B output 

    1:  ~eB3---C3
    2:  ~eB3---Z3
    3:  ~eC3---Y3
    4:  ~eL1---P3
    5:  ~eL1---S3
    6:  ~eL1---V3
    7:  ~eL2---P3
    8:  ~eL2---S3
    9:  ~eL2---V3
   10:  ~eP3---R3
   11:  ~eP3---X3
   12:  ~eP3---Z3
   13:  ~eR3---T1
   14:  ~eR3---V3
   15:  ~eR3---X3
   16:  ~eR3---Y3
   17:  ~eR3---Z3
   18:  ~eR3---b1
   19:  ~eS3---T2
   20:  ~eS3---T3
   21:  ~eS3---W3
   22:  ~eS3---X3
   23:  ~eS3---Y3
   24:  ~eS3---b2
   25:  ~eT1---V1
   26:  ~eT1---c1
   27:  ~eT2---V2
   28:  ~eT2---c2
   29:  ~eT3---W3
   30:  ~eT3---b3
   31:  ~eV1---Z1
   32:  ~eV2---Z2
   33:  ~eV3---W3
   34:  ~eV3---Z1
   35:  ~eV3---Z3
   36:  ~eV3---c1
   37:  ~eW3---Z2
   38:  ~eW3---c2
   39:  ~eX3---Y3
   40:  ~eY3---c3
   41:  ~eZ1---b1
   42:  ~eZ2---b2
   43:  ~eb3---c3
 Number of edges: 43

\end{verbatim}
\end{multicols} 
}

\dasBild{\AbbZwei}

We obtain the undirected graph in Fig.\ref{abbzwei}. 
Please notice at this stage, 
 that there are three edges $Y_3-X_3$, $S_3-W_3$ and $R_3-V_3$
 not present in the 
original graph of Fig.\ref{abbeins}. We shall not be alerted by this fact as 
the step D of FCI possibly removes further edges.\\

Let us turn to step C of FCI. We orient stepwise edges to
(see Fig.\ref{abbdrei}) :\\

{\footnotesize 
\begin{multicols}{2}
\begin{verbatim}
 FCI stage C output 
    1:  ~eB3o-oC3
    2:  ~eB3o-oZ3
    3:  ~eC3o->Y3
    4:  ~eL1o->P3
    5:  ~eL1<-oS3
    6:  ~eL1<-oV3
    7:  ~eL2o->P3
    8:  ~eL2<-oS3
    9:  ~eL2<-oV3
   10:  ~eP3<-oR3
   11:  ~eP3<->X3
   12:  ~eP3<-oZ3
   13:  ~eR3<->T1
   14:  ~eR3<->V3
   15:  ~eR3<->X3
   16:  ~eR3o->Y3
   17:  ~eR3<-oZ3
   18:  ~eR3<-ob1
   19:  ~eS3<->T2
   20:  ~eS3<-oT3
   21:  ~eS3<->W3
   22:  ~eS3o->X3
   23:  ~eS3<->Y3
   24:  ~eS3<-ob2
   25:  ~eT1<-oV1
   26:  ~eT1o-oc1
   27:  ~eT2<-oV2
   28:  ~eT2o-oc2
   29:  ~eT3o-oW3
   30:  ~eT3o-ob3
   31:  ~eV1o->Z1
   32:  ~eV2o->Z2
   33:  ~eV3o->W3
   34:  ~eV3<->Z1
   35:  ~eV3o-oZ3
   36:  ~eV3<-oc1
   37:  ~eW3<->Z2
   38:  ~eW3<-oc2
   39:  ~eX3<->Y3
   40:  ~eY3<-oc3
   41:  ~eZ1o-ob1
   42:  ~eZ2o-ob2
   43:  ~eb3o-oc3
 Number of edges: 43

\end{verbatim}
\end{multicols}
}

which is in agreement with the original graph
up to the following edges: 
$Y_3<->X_3$, $S_3<->W_3$ and $R_3<->V_3$ which are superfluous 
and $P_3<->X_3$ oriented contradictory to intention of the original graph:\\

\dasBild{\AbbDrei}

In this way we obtain the partial including path graph of 
Fig.\ref{abbdrei}.

We arrive at step D) of the algorithm. 
\newpage

\begin{verbatim}
 FCI stage D protocol (-only a part thereof)

Original network:
    Between nodes R3 and V3 D-Sep {T1,V1,X3,Z3,b1}
FCI-(1)-Derived network:
    Between nodes R3 and V3 (possible) D-Sep 
{P3,L1,L2,X3,S3,Y3,C3,c3,T2,V2,T3,W3,Z2,c2,b2,Z3,T1,V1,b1}
Edge removal:  ~eR3   V3 SepSet {T1,V1,b1}

Original network:
    Between nodes S3 and W3  D-Sep {T2,V2,T3,Y3,b2}
FCI-(1)-Derived network:
    Between nodes S3 and W3 (possible) D-Sep 
{L1,V3,L2,T2,V2,T3,X3,P3,R3,Z3,T1,V1,b1,Y3,C3,c3,b2}
Edge removal:  ~eS3   W3 SepSet {T2,V2,b2}
Original network:
    
Between nodes X3 and Y3 D-Sep {R3,T1,V1,Z3,V3,b1,S3}
FCI-(1)-Derived network:
    Between nodes X3 and Y3 (possible) D-Sep {P3,L1,L2,R3,Z3,T1,V1,b1,S3}
Original network:
    Between nodes Y3 and X3  D-Sep {C3,R3,S3,T2,V2,T3,W3,b2,c3}
FCI-(1)-Derived network:
    Between nodes Y3 and X3 (possible) D-Sep {C3,R3,S3,T2,V2,T3,b2,c3}
\end{verbatim}

{\footnotesize 
\begin{multicols}{2}
\begin{verbatim}

 FCI stage D output 

    1:  ~eB3o-oC3
    2:  ~eB3o-oZ3
    3:  ~eC3o->Y3
    4:  ~eL1o->P3
    5:  ~eL1<-oS3
    6:  ~eL1<-oV3
    7:  ~eL2o->P3
    8:  ~eL2<-oS3
    9:  ~eL2<-oV3
   10:  ~eP3<-oR3
   11:  ~eP3<->X3
   12:  ~eP3<-oZ3
   13:  ~eR3<->T1
   14:  ~eR3<->X3
   15:  ~eR3o->Y3
   16:  ~eR3<-oZ3
   17:  ~eR3<-ob1
   18:  ~eS3<->T2
   19:  ~eS3<-oT3
   20:  ~eS3o->X3
   21:  ~eS3<->Y3
   22:  ~eS3<-ob2
   23:  ~eT1<-oV1
   24:  ~eT1o-oc1
   25:  ~eT2<-oV2
   26:  ~eT2o-oc2
   27:  ~eT3o-oW3
   28:  ~eT3o-ob3
   29:  ~eV1o->Z1
   30:  ~eV2o->Z2
   31:  ~eV3o->W3
   32:  ~eV3<->Z1
   33:  ~eV3o-oZ3
   34:  ~eV3<-oc1
   35:  ~eW3<->Z2
   36:  ~eW3<-oc2
   37:  ~eX3<->Y3
   38:  ~eY3<-oc3
   39:  ~eZ1o-ob1
   40:  ~eZ2o-ob2
   41:  ~eb3o-oc3
 Number of edges: 41
\end{verbatim}
\end{multicols}
}
\newpage

Two of the unwanted edges 
$S_3<->W_3$ and $R_3<->V_3$
are removed correctly, but the third $Y_3<->X_3$
not due to nodes of D-Sep missing in Possible-D-Sep.

\dasBild{\AbbVier}

\newpage

 As a result we obtain the graph of Fig.\ref{abbvier}. 
If we apply now step E) 
of FCI  obtaining erroneous edge  $Y_3<->X_3$ and erroneous edge orientation
$P_3<->X_3$ 
indicating erroneous (conditional)  dependence of $Y_3$ on $X_3$ 
(within the original graph e.g. $\{R_3,S_3,Z_3,V_3\}$ d-separates both) 
and  conditional independence of $S_3$ and $P_3$ on $\{L_1,L_2,L_3,L_4\}$
whereas in the original network also $X_3$ is needed to d-separate both. 
We obtain also a contradictory information: the constraint
$S_3*-\underline{*X_3*}-P_3$ and at the same time edge orientations:
$S_3->X_3<->P_3$. 

Notice that failure to remove edge  $Y_3<->X_3$ is related to the sequence
of checking edges for removal. If this edge were tried first, then no error
would occur.\\

Theorem 1 is proven.
 

\section{Modification of Definition of Possible-D-Sep} 
 
\Bilder{\input FCIER.PIC}

\dasBild{\AbbSechs}

Notice that within the original definition of Possible-D-Sep 
we encountered a superfluous phrase: "Y is not a definite non-collider".
In the light of the above result
this encouraged me to assume that authors meant something different than they
actually have written. Notice that the original definition (x) of
Possible-D-Sep
makes Possible-D-Sep different from D-Sep even in fully oriented including
path graph.  Hence I assumed that the authors possibly intended to make both
Possible-D-Sep and D-sep identical for fully oriented including path graph.
Therefore I redefined Possible-D-Sep as \\
 
\begin{itemize}
\item[x'')] 
For a given partially constructed partially oriented including path graph
$\pi$, 
 V is in Possible-D-Sep(A,B) iff $V\neq A$ and there is an undirected path
from V to A such 
 that all the nodes on the path are either colliders or can be made ones by
reorientation of non-oriented edge ends, and have either A or B as their
definite successor or by reorientation of non-oriented edge-ends either A or B
can be made their definite successor. 
\end{itemize}

This definition removed the trouble that was evident with the previous
example. However, another network listed below (see Fig. \ref{abbsechs}) 
{\footnotesize 
\begin{multicols}{2}
\begin{verbatim}
NODES 
 ~np1 P_1      120  135
 ~nq1 Q_1      120  165
 ~nY1 Y_1      200  110
 ~nX1 X_1      100  190
 ~nZ1 Z_1       20  110
 ~nT1 T_1       20  190
 ~nV1 V_1        0  150
 ~nS1 S_1       80  150

 ~np2 P_2      120  235
 ~nq2 Q_2      120  265
 ~nY2 Y_2      200  210
 ~nX2 X_2      100  290
 ~nZ2 Z_2       20  210
 ~nT2 T_2       20  290
 ~nV2 V_2        0  250
 ~nS2 S_2       80  250

 ~np3 P_3      120  335
 ~nq3 Q_3      120  365
 ~nY3 Y_3      200  310
 ~nX3 X_3      100  390
 ~nZ3 Z_3       20  310
 ~nT3 T_3       20  390
 ~nV3 V_3        0  350
 ~nS3 S_3       80  350

 ~np4 P_4      120  435
 ~nq4 Q_4      120  465
 ~nY4 Y_4      200  410
 ~nX4 X_4      100  490
 ~nZ4 Z_4       20  410
 ~nT4 T_4       20  490
 ~nV4 V_4        0  450
 ~nS4 S_4       80  450

 ~np5 P_5      120  535
 ~nq5 Q_5      120  565
 ~nY5 Y_5      200  510
 ~nX5 X_5      100  590
 ~nZ5 Z_5       20  510
 ~nT5 T_5       20  590
 ~nV5 V_5        0  550
 ~nS5 S_5       80  550

 ~np6 P_6      120  635
 ~nq6 Q_6      120  665
 ~nY6 Y_6      200  610
 ~nX6 X_6      100  690
 ~nZ6 Z_6       20  610
 ~nT6 T_6       20  690
 ~nV6 V_6        0  650
 ~nS6 S_6       80  650

 ~np7 P_7      120  735
 ~nq7 Q_7      120  765
 ~nY7 Y_7      200  710
 ~nX7 X_7      100  790
 ~nZ7 Z_7       20  710
 ~nT7 T_7       20  790
 ~nV7 V_7        0  750
 ~nS7 S_7       80  750

EDGES

Sub-Network 1
    
 ~eZ1<->X1
 ~eT1<->S1 ~eS1<->Y1 ~eS1-->X1  
 ~eS1-->p1 ~eS1-->q1 
 ~eV1-->Z1 ~eV1-->T1

 ~ep1<--Z1 ~ep1<--T1 ~ep1<--V1 ~ep1<--X1
 ~eq1<--Z1 ~eq1<--T1 ~eq1<--V1 ~eq1<--X1

Sub-Network 2
    
 ~eZ2<->X2
 ~eT2<->S2 ~eS2<->Y2 ~eS2-->X2  
 ~eS2-->p2 ~eS2-->q2 
 ~eV2-->Z2 ~eV2-->T2

 ~ep2<--Z2 ~ep2<--T2 ~ep2<--V2 ~ep2<--X2
 ~eq2<--Z2 ~eq2<--T2 ~eq2<--V2 ~eq2<--X2

Sub-Network 3
    
 ~eZ3<->X3
 ~eT3<->S3 ~eS3<->Y3 ~eS3-->X3  
 ~eS3-->p3 ~eS3-->q3 
 ~eV3-->Z3 ~eV3-->T3

 ~ep3<--Z3 ~ep3<--T3 ~ep3<--V3 ~ep3<--X3
 ~eq3<--Z3 ~eq3<--T3 ~eq3<--V3 ~eq3<--X3

Sub-Network 4
    
 ~eZ4<->X4
 ~eT4<->S4 ~eS4<->Y4 ~eS4-->X4  
 ~eS4-->p4 ~eS4-->q4 
 ~eV4-->Z4 ~eV4-->T4

 ~ep4<--Z4 ~ep4<--T4 ~ep4<--V4 ~ep4<--X4
 ~eq4<--Z4 ~eq4<--T4 ~eq4<--V4 ~eq4<--X4

Sub-Network 5
    
 ~eZ5<->X5
 ~eT5<->S5 ~eS5<->Y5 ~eS5-->X5  
 ~eS5-->p5 ~eS5-->q5 
 ~eV5-->Z5 ~eV5-->T5

 ~ep5<--Z5 ~ep5<--T5 ~ep5<--V5 ~ep5<--X5
 ~eq5<--Z5 ~eq5<--T5 ~eq5<--V5 ~eq5<--X5

Sub-Network 6
    
 ~eZ6<->X6
 ~eT6<->S6 ~eS6<->Y6 ~eS6-->X6  
 ~eS6-->p6 ~eS6-->q6 
 ~eV6-->Z6 ~eV6-->T6

 ~ep6<--Z6 ~ep6<--T6 ~ep6<--V6 ~ep6<--X6
 ~eq6<--Z6 ~eq6<--T6 ~eq6<--V6 ~eq6<--X6

Sub-Network 7
    
 ~eZ7<->X7
 ~eT7<->S7 ~eS7<->Y7 ~eS7-->X7  
 ~eS7-->p7 ~eS7-->q7 
 ~eV7-->Z7 ~eV7-->T7

 ~ep7<--Z7 ~ep7<--T7 ~ep7<--V7 ~ep7<--X7
 ~eq7<--Z7 ~eq7<--T7 ~eq7<--V7 ~eq7<--X7

Internetwork connections  
Giving nodes X, p, and q to other networks
 ~eZ3-->X1 ~ep1-->Y3  ~eT4-->X1 ~eq1-->Y4
 ~eZ4-->X2 ~ep2-->Y4  ~eT5-->X2 ~eq2-->Y5
 ~eZ5-->X3 ~ep3-->Y5  ~eT6-->X3 ~eq3-->Y6
 ~eZ6-->X4 ~ep4-->Y6  ~eT7-->X4 ~eq4-->Y7
 ~eZ7-->X5 ~ep5-->Y7  ~eT1-->X5 ~eq5-->Y1
 ~eZ1-->X6 ~ep6-->Y1  ~eT2-->X6 ~eq6-->Y2
 ~eZ2-->X7 ~ep7-->Y2  ~eT3-->X7 ~eq7-->Y3

\end{verbatim}
\end{multicols}
}
\dasBild{\AbbSieben}

\noindent
 can be constructed which will make the FCI algorithm
with this definition also
 invalid,  as visible from
Figs.\ref{abbsieben}-\ref{abbneun} - see superfluous edges $X_i<->Y_i$ in
Fig.\ref{abbneun}, also
incorrect edge orientations of edges $X_i<->P_i$, $X_i<->Q_i$.
Notice, that this network would have made no trouble to the previous version
of Possible-D-Sep 

\dasBild{\AbbAcht}
\dasBild{\AbbNeun}

\section{Modification of FCI Algorithm}
 
Let us suppose the following 
intended meaning of Possible-D-Sep. It should always be superset of D-Sep
and errors in orientation of intrinsic edges resulting from presence of
superfluous edges in a partially including paths graphs as well as presence
of superfluous edges themselves should not remove any D-Sep nodes from
Possible-D-Sep. In this case, however, both the definition of Possible-D-Sep
and the FCI algorithm itself need to be modified. They should not rely on 
presence of arrows at edge ends because they are a property which is nether
truth-preserving nor falsehood-preserving on removal of superfluous
edges.
The truth-preserving property in that case is the presence of local constraint
information $A*-\underline{*B*}-*C$.  Therefore let us redefine the notion
of Possible-D-Sep as follows:\\
 
\begin{itemize}
\item[x''')] 
For a given partially constructed partially oriented including path graph
$\pi$, {\bf Possible-D-Sep}(A,B) is defined as follows: 
If $A\neq B$, V is in  {\bf Possible-D-Sep}(A,B) in $\pi$ if and only if
$V\neq A$, and there is an undirected path U between A and V in $\pi$ such
that for every subpath $<X,Y,Z>$ of U 
we have {\em no } local constraint
information $X*-\underline{*Y*}-*Z$ in $\pi$.  
\end{itemize}

Furthermore stage C of original FCI algorithm has to be replaced by the
following prescription:

\begin{description}
\item[C')] Let F be the undirected graph resulting from step B). Orient each 
edge as $o-o$. For each triple of vertices A,B,C such that the pair A,B and 
the pair B,C are each adjacent in F, but the pair A,C are not adjacent in F, 
orient $A*-*B*-*C$ as  $A*-\underline{*B*}-*C$ if and only if B is in
Sepset(A,C).
\end{description}

Summarizing, both definition of Possible-D-Sep and the stage C) of FCI have
to consume or produce resp. local constraint information instead of
head-to-head edge orientation ("colliders"). This in my opinion corrects the
algorithm completely and correctly. To prove this claim briefly, let us first
turn to relationship between Possible-D-Sep and D-Sep in a fully oriented
intrinsic including path graph.  Obviously, any node in D-Sep will also
belong to Possible-D-Sep as the path out of collider nodes in D-Sep excludes
any local constraint information  $X*-\underline{*Y*}-*Z$ for any three
subsequent nodes on this path. Let us now consider a partially constructed
partially oriented including path graph, if $X-Y$ and $Y-Z$ are intrinsic
connections then  $X*-\underline{*Y*}-*Z$ information in the partially
constructed partially including path graph would immediately imply that
$X*-\underline{*Y*}-*Z$  holds also in the intrinsic underlying including path
graph - as this relies not on graph properties but solely on conditional
independence property. Hence nothing like this appears on an intrinsic
collider path, hence Possible-D-Sep will always contain D-Sep. Notice that in
a partially constructed partial including path graph superfluous 
local
constraints information  $A*-\underline{*B*}-*C$ may occur in case that for
example edge $A-B$ is superfluous one. But this does not disturb the algorithm
in any way as does not influence relationship between Possible-D-Sep and
D-Sep. In some sense it may be considered as a correct information, because an
intrinsic edge can never meet head-to-head with a non-existent edge. 

\section{Discussion}

The paper demonstrates by example  that the FCI algorithm as it stands in
\cite{Spirtes:93}, is not correct. To repair it, it is necessary either to
drop  step 
C) spoiling the whole algorithm altogether, 
or to change both the definition of Possible-D-Sep and step C. 
This is 
because of the philosophy of FCI: it is meant to remove first as much edges 
as possible using only direct neigbbours of a node (just taking advantage of 
earlier edge removals in the process), and then to take into account those 
nodes, which are not neighbouring but influence dependence relationship 
between the nodes which is done in step D). Step C) was intended as a way to 
bind set of  potential candidates for dependency considerations of step D), 
but is just demonstrated to be wrong. Obviously, it can be removed without 
violating the philosophy of the algorithm, while re-establishing algorithm's 
correctness. 
FCI algorithm modified by removal of step C), however, would not be too 
beneficial compared to the primary CI algorithm but for really sparse 
networks. And hence, like CI, would be rarely applicable to networks larger 
than 
a few nodes. \\
Therefore the  alternative presented in section 6 seems to be reasonable. It
is, however, more
space consuming - due to the necessity of maintaining local constraint list.
However, when starting stage C of CI (as required at the end of FCI), the
local constraint list could be retained to save re-calculations.\\

We could instead take the policy - as an alternative to section 6 approach -
that we would rerun steps C and D whenever an edge has
been removed in stage D. This seems however not to be a time-efficient
solution. 

Still another alternative could be to postpone removal of an edge detected
as superfluous in Stage D until the test D is completed for all the other
edges. This approach  requires only one logical cell of storage for each edge
(to notice whether or not the edge has to be removed after completion of stage
D). However, as with the original algorithm, the orientation information for
edges would then have to be dropped and later re-calculated in stage C of
CI.

Two facts about the sample network used in the proof of Theorem 1 cannot be 
overseen: the network as such is rather a big one  and this network is
artificially constructed, and therefore may seldom occur in practice.
However, from the point of view of statistics it is not negligible that an
algorithm makes systematic errors beside random ones. 

The lesson to be learned from failure of FCI is that one should be very 
careful if a network structure algorithm runs at risk of introducing 
non-existent links between nodes, especially if it is based on local criteria 
like FCI and CI.

The result of this paper has consequences for the validity of the theory 
developed in Chapter 6 of \cite{Spirtes:93}. Our result means directly that 
the Theorem 6.4 of  \cite{Spirtes:93} is wrong, and all the claims derived 
from it should be at least reconsidered.

\section{Conclusions}

In this paper non-suitability of the FCI algorithm of Spirtes et al,
as it stands in 
\cite{Spirtes:93}, for recovery of belief networks from data under causal 
insufficiency has been proven by example. It must be acknowledged that the 
 size of the demonstration network is considerable, and hence under practical
settings under which this type of 
algorithms is applied, should seldom have a chance to emerge. However, 
 users of the algorithm should be aware of possibilities of occasional 
failures 
built into the philosophy of the algorithm. 

The only ad hoc possibility of repair for FCI is to drop its step C) 
altogether, but then improvement over CI of \cite{Spirtes:93} is only for very
sparse networks, and CI is known to be feasible for networks with mean and 
large number of nodes. 

A more elaborate repair method has been proposed which changes the
definition of Possible-D-Sep and stage C of FCI algorithm.

Further research efforts are necessary to establish 
other derivatives 
 of the CI algorithm which would be computationally feasible but not lead to 
incorrect  results by definition.


\end{document}